\documentclass[11pt]{article}

\usepackage[final]{acl}

\usepackage{times}
\usepackage{latexsym}

\usepackage[T1]{fontenc}

\usepackage[utf8]{inputenc}

\usepackage{microtype}

\usepackage{inconsolata}

\usepackage{graphicx}

\usepackage{times}
\usepackage{booktabs}
\usepackage{soul}
\usepackage{caption}
\usepackage{amsmath}
\usepackage{amssymb}
\usepackage{amsthm}
\usepackage{algorithm}
\usepackage{algorithmic}
\usepackage{adjustbox}
\usepackage{multirow}
\usepackage{longtable}
\usepackage{wrapfig}
\usepackage{subcaption}
\usepackage{siunitx}

\definecolor{formula_color}{HTML}{8CC5DF}

\newcommand{\datasetname}{FormulaReasoning}
\newcommand{\red}[1]{\textcolor{red}{#1}}
\newcommand{\green}[1]{\textcolor{green}{#1}}
\newcommand{\revision}[1]{#1}

\title{\datasetname: A Dataset for Formula-Based Numerical Reasoning}

\author{Xiao Li\textsuperscript{*}\and Bolin Zhu\thanks{These authors contributed equally to this work.}\and Kaiwen Shi\\ \textbf{Sichen Liu\and Yin Zhu\and Yiwei Liu\and Gong Cheng}\thanks{Corresponding author} \\
  State Key Laboratory for Novel Software Technology, Nanjing University, Nanjing, China \\
  \texttt{\{xiaoli.nju,bolinzhu\}@smail.nju.edu.cn} \\ 
  \texttt{\{kaiwenshi,sichenliu,yinzhu,ywliu\}@smail.nju.edu.cn} \\ \texttt{gcheng@nju.edu.cn} \\
}

\begin{document}
\maketitle
\begin{abstract}
The application of physics formulas is a fundamental human capability in numerical reasoning. While existing datasets often rely on implicit mathematical knowledge, they rarely explicitate the underlying formulas. To address this, we introduce \datasetname, a new benchmark for formula-based numerical reasoning comprising 5,324 questions requiring calculations grounded in external physics principles. We provide high-quality, fine-grained annotations in English and Chinese—including formula structures, parameter names, symbols, values, and units—curated through manual effort and LLM-assisted validation. Additionally, we provide a consolidated formula database as an external knowledge source. To further challenge model performance, we develop an extended version of the dataset by coupling multiple questions. We evaluate various architectural and methodological frameworks, including retrieval-augmented methods, modular reasoning (formula generation, parameter extraction, and calculation), and preference-based optimization. Our analysis identifies critical challenges in formula-based reasoning, highlighting significant opportunities for future methodological advancement.
\end{abstract}

\section{Introduction}
\label{sec:intro}

Numerical reasoning constitutes one of the significant forms within natural language reasoning~\citep{Frieder2023MathematicalCO}. The study of numerical reasoning has seen substantial progress in recent years, driven largely by the development of LLMs~\citep{gpt4,llama2,qwen2} and specialized datasets~\citep{wang-etal-2017-deep,dua-etal-2019-drop,amini-etal-2019-mathqa,gsm8k}. %
Current datasets for numerical reasoning typically include simple commonsense numerical questions that under-reflect the complexity of real-world problems.
These datasets either do not provide process supervision information, or the provided reasoning steps are essentially incomplete, as they often rely on implicit commonsense knowledge but not explicit knowledge guiding the reasoning process.
This issue becomes particularly evident when LLMs experience hallucination~\citep{Frieder2023MathematicalCO,Bang2023AMM}, especially in the absence of clear knowledge guidance.
Consequently, one might ask \emph{``What explicit knowledge could be used to guide a numerical reasoning process?''}. Formulas exactly represent such knowledge that has been largely overlooked in previous research but is frequently utilized in real-life applications.

\begin{figure*}[t]
    \centering
    \includegraphics[width=0.95\linewidth]{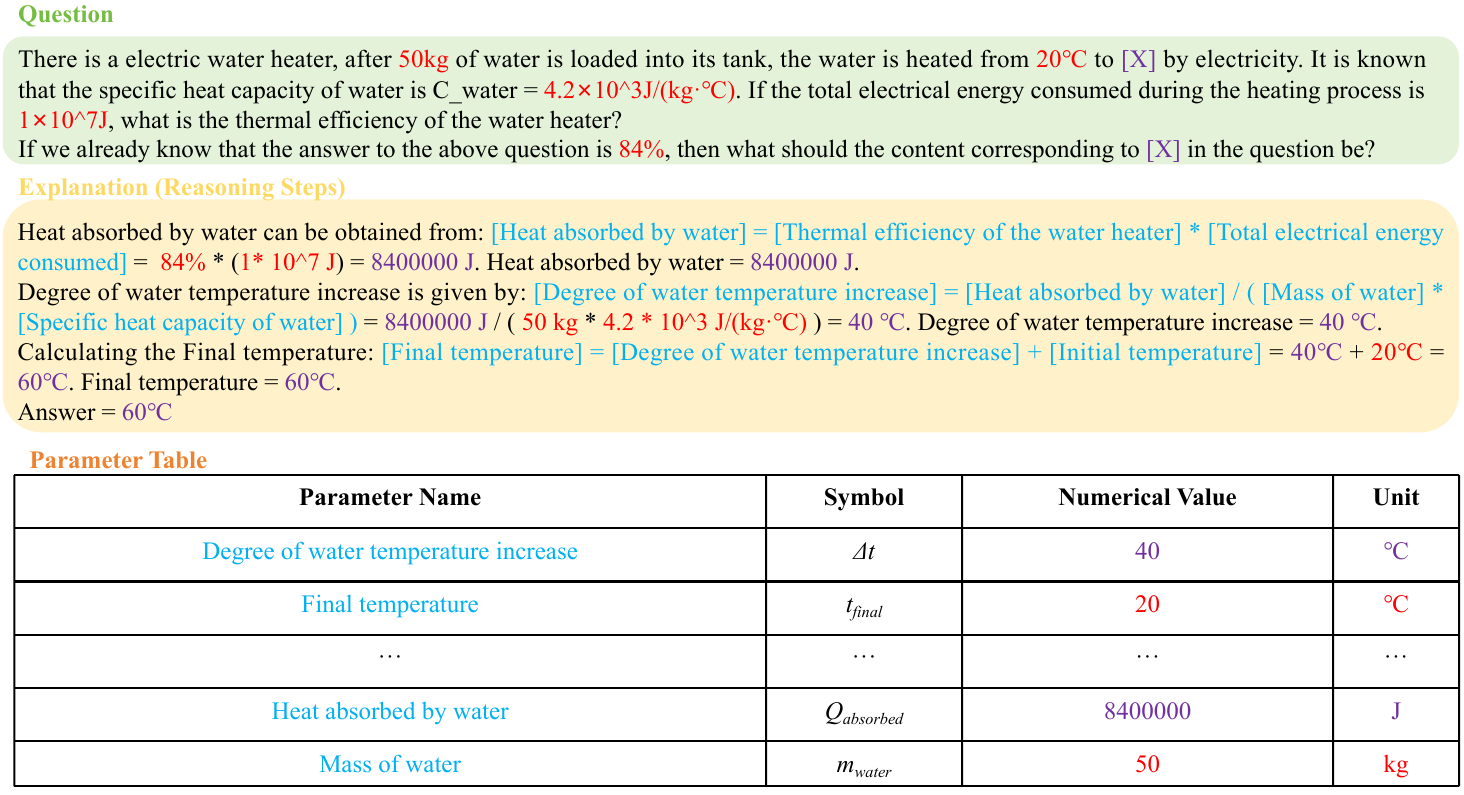}
    \caption{An example from \datasetname. Numerical values with units given in the question and obtained from intermediate steps are highlighted in red and purple, respectively. Formulas and their elements are in blue.}
    \label{fig:example}
\end{figure*}

\begin{table*}[t]
\small
\centering
\caption{Statistics of Math23K-F, MAWPS-F, GSM8K, MATH, and our \datasetname.}
\adjustbox{width=0.8\linewidth}{
\begin{tabular}{lrrrrr} 
\toprule 
Dataset & Math23K-F & MAWPS-F & GSM8K & MATH & \datasetname \\
\midrule 
    \# questions & 23,162 & 2,373 & 8,792 & 12,500 & 5,324 \\
    \# questions requiring formulas & 7,750 & 911 & N/A & N/A & 5,324  \\
    \# formulas (and variants) & 51~(131) & 18~(46) & 0~(0) & 0~(0) & 272~(845) \\
    Avg. \# reasoning steps & 1.16 & 1.01 & 3.59 & Not Provided & 2.36  \\
\bottomrule
\end{tabular}
}
\label{table:datasetstatistics}
\end{table*}

\paragraph{Limitations of existing datasets}
Take a question from the GSM8K dataset~\citep{gsm8k} as an example: ``A robe takes 2 bolts of blue fiber and half that much white fiber. How many bolts in total does it take?'' This question implicitly uses \emph{simple commonsense mathematical knowledge}~(e.g., half means dividing by~2) to solve without requiring complex domain-specific formular knowledge for numerical reasoning. 
Recently, Liu \emph{et al.}~\cite{Liu2023GuidingMR} constructed two formula-based datasets, Math23K-F and MAWPS-F. Although 33\% and 38\% of the questions in these datasets, respectively, require the use of formulas, they still mainly consist of simple commonsense formulas~(e.g., total\_amount = unit\_amount $\times$ total\_number).
Currently, \emph{there is a lack of datasets where questions require domain-specific formulas to guide complex numerical reasoning}, such as the physics formula used to calculate the heat absorption of an object. 
Related work is in Appendix~\ref{sec:related_work}.

\paragraph{Our work}
To fill this gap, we introduce \datasetname, a dataset for numerical reasoning requiring domain-specific (physics) formulas. We provide fine-grained annotations (Figure~\ref{fig:example}), enabling effective differentiation of model capabilities, especially for smaller models.
Table~\ref{table:datasetstatistics} compares \datasetname\ with existing datasets. Compared to Math23K-F and MAWPS-F, \datasetname\ contains \emph{more diverse formulas}~(272 vs. 18--51). Moreover, the \emph{higher average number of reasoning steps}~(2.36 vs. 1.01--1.16) provides a more comprehensive evaluation framework for understanding how different model architectures handle multi-step formula-based reasoning.

Specifically, we collected questions requiring formula-based numerical reasoning from junior high school physics exams. With combined efforts of manual annotation and LLM assistance, we annotated each question with an explanation text that provides \emph{normalized reasoning steps with relevant formulas} (including formula structures, parameter names, symbols, numerical values, and units) and a final answer. We also built a \emph{consolidated formula database} that functions as an external knowledge base and can be used by retrieval-based/augmented systems. 
The original questions and annotations are in Chinese. We used an LLM to translate them into high-quality English, forming a \emph{bilingual dataset}.

We evaluated \datasetname\ using various reasoning paradigms: standard LLMs (4B to $>$100B parameters), fine-tuned models with CoT supervision, data augmentation, and RAG. We also derived preference data for DPO. Our analysis reveals significant differences in how approaches handle formula-based reasoning.

To further challenge the reasoning capabilities of models, we constructed \datasetname+. This extended version coupled each question with another question and reformulated direct calculation as solving a system of equations, significantly increasing the difficulty of numerical reasoning.

\paragraph{Broader impact} \revision{It is worth noting that while \datasetname\ is constructed within the physics domain, the underlying framework is \emph{domain-agnostic}. By abstracting formulas into an operator-parameter structure, our dataset construction pipeline can be readily extended to other disciplines requiring structured knowledge, such as finance~(e.g., compound interest formulas), chemistry~(e.g., stoichiometry), and geometry. Furthermore, beyond serving as a benchmark, \datasetname\ offers significant research value for \emph{step-wise reasoning}, \emph{structured formula retrieval}, and \emph{preference optimization (DPO)}, providing rich resources for advancing neuro-symbolic reasoning.}

Our \textbf{contributions} are summarized as follows.
\begin{itemize}
    \item We construct a formula-based numerical reasoning dataset with fine-grained annotations for each question.
    It can be applied to evaluate knowledge-guided reasoning capabilities. Its normalized reasoning steps are useful for the emerging study of step-supervised reasoning.
    \item We perform an evaluation of LLMs of various sizes, RAG, fine-tuned small models, and DPO methods based on derived preference data. Our experimental results establish a solid baseline for future research and also indicate that the performance disparity across model scales presents new research opportunities for efficient formula-based reasoning.
\end{itemize}

\datasetname\ is available from anonymous GitHub \url{https://github.com/nju-websoft/FormulaReasoning} under Apache 2.0.

\section{Dataset Construction}
\label{sec:dataset}

We collected questions from Chinese junior high school physics exams. We invited five graduate students to act as annotators, all of whom hold a bachelor's degree in science and engineering. For all questions, we normalized their explanations. Each question was annotated with normalized reasoning steps in natural language and a tabular representation of the steps using formulas, including the numerical values and units for all the parameters in the formulas. This process involved a combination of manual annotation and the assistance of an LLM\footnote{During dataset construction, we consistently accessed Qwen-max as our LLM via its \href{https://help.aliyun.com/zh/dashscope/developer-reference/quick-start}{API}. } to improve the efficiency of annotation. We compiled all the formulas and merged those that express the same meaning to create a consolidated formula database. Finally, we altered the questions to avoid label leakage and translated the dataset into English.
In the following, we elaborate this process to construct \datasetname.

\subsection{Preprocessing}
We crawled 18,433 junior high school physics exam questions in China from 2015 to 2024 from public sources, including only those with free-text answers and excluding multiple-choice and true/false questions. Each raw question contains a \emph{question text} and an \emph{explanation text that includes reasoning steps}. We eliminated questions requiring diagrams.

We filtered the questions by identifying the presence of numerical values within the explanation and confirming that the final answer was numerical. Using regular expressions, we extracted the \emph{final numerical answer} including its unit from the explanation. We found that for 487 questions, our regular expressions did not return results, so we manually extracted their answers from the explanation text.
After this preprocessing, we compiled an initial dataset comprising 6,306 questions.

\begin{table}[t]
\centering
\small
\caption{Original explanation and explanation with normalized formulas~(highlighted in blue).}
\begin{tabular}{|p{\linewidth}|}
    \hline
    \textbf{Original explanation:} \\
    The change in water temperature is 60 - 20 = 40 °C. Therefore, the heat absorbed by the water is Q\_\{absorbed\}=50 kg $\times$ 4.2 $\times 10^3$ J/(kg·°C) $\times$ 40 °C = 8.4 $\times 10^6$ J. Given that the total electrical energy consumed in the heating process is $1\times 10^7$ J, the thermal efficiency of the water heater can be calculated using the formula for the efficiency of a heat engine: $\eta$ = Q\_\{absorbed\}\}/W\_\{total\}$\times 100$\% = ($8.4\times 10^6$ J)/($1.0\times 10^7$ J)$\times 100$\% = $84$\%. Answer: If it is known that the total electrical energy consumed during the heating process is $1\times 10^7$, the thermal efficiency of the water heater is $84$\%.\\ \\
    \textbf{Explanation with normalized formulas:} \\
    1. Calculating the temperature increase in water: \textcolor{formula_color}{[Degree of water temperature increase] = [Final temperature] - [Initial temperature]} = 60 ℃ - 20 ℃ = 40 ℃. The degree of water temperature increase = 40 ℃. \\ 
    2. Calculating the heat absorbed by water: \textcolor{formula_color}{[Heat absorbed by water] = [Mass of water] $\times$ [Specific heat capacity of water] $\times$ [Degree of water temperature increase]} = 50 kg $\times$ 4.2 $\times$ $10^3$ J/(kg·℃) $\times$ 40 ℃ = 8400000 J.  The heat absorbed by water = 8400000 J. \\ 
    3. The thermal efficiency of the water heater can be obtained from: \textcolor{formula_color}{[Thermal efficiency of the water heater] = [Heat absorbed by water] / [Total electrical energy consumed] $\times$ 100\%} = 8400000 J / ($1\times 10^7$ J) * 100\% = 84\%. The thermal efficiency of the water heater = 84\%. \\ 
    Answer = 84\% \\
    \hline
\end{tabular}
\label{table:explanation_cmp}
\end{table}

\subsection{Formula Normalization}
\label{sec:formula_normalization}

The reasoning steps in the raw explanation text lacked a normalized format and were expressed casually. Some formulas mixed parameter names~(e.g., ``mass of water'') with symbols~(e.g., ``$m_{water}$''), while others simply provided calculations over numerical values without parameter names or symbols.
To ensure that all explanations adopted a common form of formulas, we performed normalization, as illustrated in Table~\ref{table:explanation_cmp}.
We aimed to \emph{identify and normalize the formulas in the original explanations} and then \emph{verify and unify their formats}. Manually carrying out such tasks would require a significant effort. 
However, since the process is not open-ended, but rather structured and verifiable,
we could automatically, e.g. \emph{using an LLM}, extract formulas from a normalized explanation, symbolicly calculate each step, and compare the result with the given answer to ensure the accuracy of our normalization.

Specifically, to improve the efficiency of annotation, we adopted a \emph{coarse-to-fine annotation process} with the help of an LLM.
We first prompted the LLM to revise the reasoning steps (in particular, the formulas) in the explanation text into a normalized form. Then, we prompted the LLM to correct minor errors within the normalized explanations, including formatting issues in formula annotations and inaccuracies in the parameters used during calculation. In the following, we elaborate this two-stage process. More details, including the prompts, are provided in Appendix~\ref{appendix:prompts_formula_standardization}.

\paragraph{Coarse-grained annotation}
We introduced each question with its original explanation and answer to guide the LLM through a few-shot prompt to normalize the explanation.
We observed that the quality of the normalized explanations was generally satisfactory.
We also required the LLM to present the symbol, numerical value, and unit of each parameter in formulas in the form of a table, as illustrated in Figure~\ref{fig:example}.

\paragraph{Fine-grained annotation}
We checked the correctness of the formula format in the explanations by rules, including whether there were omissions in parameter names, symbols, or units, and these minor issues were correctable.
\revision{To assess the correctness of each normalized explanation, we extracted formulas from the explanation and calculated an answer using the Numbat calculator\footnote{\href{https://numbat.dev}{https://numbat.dev}. Numbat is designed for scientific computation with support for physical units.}. We achieved a formula normalization accuracy of 85.95\% through this programmatic verification.
We used few-shot prompts to correct LLM errors and removed poor-quality questions, such as those missing reasoning steps, as illustrated in Appendix~\ref{appendix:examples_of_deleted_questions}. After that, our dataset contains 5,420 questions remaining.
To further improve data quality, we performed a rigorous human audit on a critical subset of the data. Specifically, we manually reviewed and corrected 363 questions~(approximately 45.4\% of the test set; see Section~\ref{sec:dataset_split} for dataset split), which primarily consisted of questions where the model predictions deviated from the ground truth.}

\subsection{Formula Database Construction}

Our next step was to \emph{construct a consolidated formula database for the entire dataset}. Parameters in the same formula can be expressed differently in various problem contexts. For example, the two formulas ``[weight of water] = [mass of water] * [gravitational acceleration]'' and ``[weight] = [mass] * [gravitational acceleration]'' both calculate the weight of an object.
It would be helpful to standardize these formulas for use with LLMs, particularly in methods such as RAG~\cite{tran2024rare}, step supervision~\cite{zhang2024llama}, and tree-based search~\cite{zhao2024marco},
to avoid long-tail distribution issues caused by treating different formula expressions as different formulas.

\begin{table}[t]
\small
\centering
\caption{Changes in the number of formulas after each merging step.}
\begin{tabular}{lr}
    \toprule
    Step & \# Formulas \\
    \midrule
    Before merging & 12,906  \\
    After symbolic rules based merging & 1,163  \\
    After semantics based merging &  439 \\
    After manual review and error correction & 272  \\
    \bottomrule
\end{tabular}
\label{table:formula_num_changes}
\end{table}

We divided the construction process of a consolidated formula database into three steps: 1) Merge formulas through symbolic rules. 2) Merge formulas through a semantics-based method. 3) Manually review and correct errors.
In Table~\ref{table:formula_num_changes}, we present the initial number of formulas and the remaining number of formulas after each step.

\paragraph{Symbolic rules based merging} 
This was achieved by \emph{comparing formula structures and symbols}.
Consider the following as an example of judging whether two formulas have the same structure. The three formulas ``$f_1:~a_1=(b_1+c_1)/d_1$'', ``$f_2: a_2=(b_2+c_2)/d_2$'', and ``$f_3: b_1=a_1*d_1-c_1$'' have the same structure because $f_2$ can be derived from $f_1$ by renaming parameters, and $f_3$ can be obtained from $f_1$ by transformation.
Moreover, in physics, physical quantities are conventionally represented by specific symbols. For example, the mass of an object is often denoted by ``$m$'' and the density of an object is frequently represented by ``$\rho$''. Subscripts are then used to specify which specific object a physical quantity refers to, such as ``$\rho_{water}$'' for the density of water.
Therefore, for each pair of formulas, we first computed all possible transformations of each formula to obtain a set of all its variants. Then, we compared the formula structures in the two sets to determine if the two formulas shared a structure. If so, we checked whether their symbols, with subscripts removed, were identical. If so, we considered these two formulas to be mergeable. When merging, we retained the parameter with the shorter name of the two. 
After merging based on such symbolic rules, we reduced the number of formulas in the formula database from 12,906 to 1,163.

\paragraph{Semantics based merging} 
In the previous step, the semantic information in the parameter names was not used. This led us to \emph{perform merges grounded on the semantics of parameter names}. 
For example, two formulas that were not merged during the symbolic merging stage, ``[density] = [mass] / [volume]'' and ``[density of water] = [mass of water] / [volume of water]'', should actually be merged. We identified such mergeable formulas based on the semantic information in the parameter names, e.g., ``density'' and ``density of water'' are semantically similar.
Specifically, for formulas with identical structures, we tokenized\footnote{We used jieba: \href{https://github.com/fxsjy/jieba}{https://github.com/fxsjy/jieba}.} each pair of their corresponding parameter names into two sets of words.
When the two sets overlapped, the two parameters were considered to have a semantic connection and the two formulas became candidates for merging. Using this approach, we identified a set of pairs of potentially mergeable formulas and then consulted the LLM for a detailed evaluation of each pair. The prompt is provided in Appendix~\ref{appendix:semantic_based_merging}. After this step, the number of formulas in the formula database was reduced from 1,163 to 439.

\paragraph{Manual review and error correction} 
Upon completing the aforementioned merging process, we manually inspected the accuracy of the results, rectified the instances where errors occurred during merging, and manually merged formulas that were overlooked by the LLM. In this process, two human volunteers cross-validated the results of manual review and correction. Finally, we obtained a consolidated formula database consisting of 272 distinct formulas.

\subsection{Question Alteration}
\label{sec:question_alteration}

The original questions in \datasetname\ were collected from publicly accessible online sources. Considering that LLMs were typically trained on massive-scale corpora that potentially encompassed these publicly available resources, there existed a non-negligible risk of data contamination and label leakage. To mitigate this potential bias and construct a more rigorous benchmark, we implemented systematic strategies for question alteration. Specifically, we employed \emph{inverse transformation}~\citep{you-etal-2024-mumath} and \emph{parameter noise injection}~\citep{wei-zou-2019-eda} to generate a modified version for each original question while preserving its fundamental reasoning nature.

\paragraph{Inverse transformation}
We replaced a random parameter in the original question with a placeholder ``[X]'', and added the original answer to the question. The new question then asked for the value of the masked parameter ``[X]''. Due to the various expressions of parameters, some could not be easily found and replaced in the question by string matching, e.g., ``$4.2\times10^{6} J/h$'' and ``$1\times10^{-3} dm^{3}$'' could be written as ``$4.2$×$10^{6}$ J per hour'’ and ``1L'', respectively. We had to exclude such questions.

\paragraph{Parameter noise injection}
We randomly selected a parameter from the entire set of parameters in all questions and appended both its parameter name and its numerical value to the question as a distractor. To ensure that the added parameter would not affect the solvability of the question~(e.g., not introducing two conflicting values for the same physical quantity), we ensured that the added parameter was different from all the parameters involved in answering the original question.

To verify that the modified questions were solvable, we used SymPy\footnote{\href{https://www.sympy.org/}{https://www.sympy.org/}} to solve the equations. During the verification process, some equations could not be solved (e.g., SymPy cannot handle ``\%'' and ``1000kg/t''), so we removed these questions. The final number of the remaining questions is~5,324.

\subsection{\revision{English Version of \datasetname}}

LLMs translated the Chinese questions, explanations, and formulas into English using prompts from Appendix~\ref{appendix:dataset_translation_prompt}. All components, including parameter names and reasoning steps, were translated.

To assess translation quality, we compared the perplexity of LLM and manual translations for 50 sampled questions using the Qwen2.5-14B-Instruct model. The LLM scored \num{6.28} compared to \num{7.26} for human versions, demonstrating reasonable fluency. Additionally, a manual review of these 50 questions found no translation errors that would influence consistency or question-answering accuracy.

\subsection{\revision{Question Coupling}}
To further elevate the reasoning complexity, we introduced an extended version \datasetname+ that tightly coupled the original questions with a synthesized auxiliary question and reformulated direct calculation as solving a system of equations.
First, we selected a mutable parameter in the original question to mask as $[X]$ and denoted the original answer as $[Y]$.
Second, we constructed a linear expression $E(X, Y)$ (e.g., $c_1 Y + c_2 X$) ensuring a positive result and magnitude alignment between $X$ and $Y$ by adjusting coefficients.
Third, we calculated the value $v = E(X, Y)$ and prompted an LLM to generate an independent physics question ($Q_{aux}$) whose answer is $v$.
Finally, we formulated a hybrid question that requires first solving~$Q_{aux}$ to obtain~$v$ and then solving a system of equation in~$X$ and~$Y$ where one equation represents the original question and the other is $v=E(X,Y)$. After removing the questions with failed coupling, the final \datasetname+ contains 4,392 extended questions. It is worth noting that Question Coupling and Question Alteration (Section~\ref{sec:question_alteration}) are parallel processes, both derived from the same original questions.
For the detailed coupling process, please refer to Appendix~\ref{appendix:question_coupling}. 

\section{Experimental Setup}
\label{sec:experiments_setup}

To study how methodological choices affect formula-based reasoning in \datasetname, we analyzed prompting, fine-tuning, and retrieval-augmented methods. We developed two specific approaches: one decomposing reasoning into formula generation and calculation steps, and another utilizing data augmentation. We also explored preference learning for refinement. This setup enables a comprehensive comparison of architectural and methodological impacts on performance.

\subsection{Dataset Split}
\label{sec:dataset_split}
We divided \datasetname\ into three subsets: training, \emph{HoF}~(Homologous Formulas) test, and \emph{HeF}~(Heterologous Formulas) test, comprising 4,524, 413, and 387 questions, respectively. 
Inheriting the same split from \datasetname, \datasetname+ contains 3,608 training questions, with the \emph{HoF} test and \emph{HeF} test sets containing 406 and 378 questions, respectively.
All formulas in the HoF test set appeared in the training set, while in the HeF test set, each question required at least one formula not seen in the training set. This division was to evaluate the generalizability of fine-tuned models on new formulas.

\subsection{Evaluated Methods}
\paragraph{Human Performance}
\revision{We recruited 108 students from a high school, each student being assigned 7--8 questions. Each student was given 40 minutes to complete these questions. These questions were used as part of their in-class exercises and, at the end, each student received a gift. The final statistics were collected to assess human performance, which was consented to by all students.}

\paragraph{LLMs}
Following~\cite{cot_zero_shot}, we incorporated ``Let's think step by step'' (i.e. \textbf{CoT}) into a zero-shot prompt to guide LLMs in generating reasoning steps. 
The prompt is in Appendix~\ref{appendix:zero_few_shot_prompt_example}.

We conducted experiments across a diverse spectrum of models including both cost-friendly small-scale models and high-cost LLMs. The list of evaluated models is provided in Appendix~\ref{appendix:evaluated_llms}.
This comprehensive evaluation allowed us to assess capabilities across different model scales while identifying persistent challenges in formula reasoning.

We also compared CoT with \textbf{Program of Thought~(PoT)}~\cite{pot}. In PoT, we used a Python interpreter to execute the code and obtain an answer. 

\paragraph{Formula Retriever}
We trained a formula retriever on the training set. Specifically, we encoded each question using the Chinese-BERT-wwm-base model~\citep{devlin-etal-2019-bert,cuietal2021pretrain} to obtain the CLS vector of the question. 
The formulas in the formula database were not encoded in this way, as they contained structured information so that two textually similar formulas could have entirely different meanings~(e.g., $z_1 = x \times y$  and $z_2 = x / y$ ). Therefore, each formula was instead represented by a randomly initialized vector that was updated during training.
We calculated the cosine score between the question vector and the formula vector. The retriever was then trained with in-batch negatives and contrastive learning loss~\cite{gao-etal-2021-simcse}. During inference, for each question in the HoF test, we retrieved the top five formulas with the highest scores and included them in the prompt to augment LLM generation.
More details are provided in Appendix~\ref{appendix:formula_retriever}.

\paragraph{Supervised Fine-Tuned Models}
\label{sec:sft_models}
We found that directly prompting models having fewer than 8B parameters did not produce satisfactory results, so we performed supervised fine-tuning of small models (Qwen2.5-Math-7B-Instruct and Qwen2.5-Math-1.5B-Instruct). Unlike large models, small models struggled with numerical extraction and calculation. To enhance their reasoning capabilities, we developed: (1) \textbf{CoT-Supervised Fine-Tuning~(CoT-SFT)}, a three-step process where the model generates relevant formulas, extracts parameter values and units from questions, and uses Numbat for final calculation; and (2) \textbf{Data Augmentation~(DA)}, using Qwen-max to generate and verify new training examples with correct reasoning steps (details in Appendix~\ref{appendix:data_augmentation}).

\paragraph{Direct Preference Optimization~(DPO)}
We used the preference data derived in Appendix~\ref{DPO_construction} to experiment on DPO.
In the generation process, we used DeepSeek-R1-Distill-Qwen-7B as the generation model, and used Qwen2.5-Math-PRM-7B as the process reward model (PRM).

The implementation details are provided in Appendix~\ref{appendix:implementation_details}.

\subsection{Evaluation Metrics}
We utilized Numbat to evaluate the prediction generated by each model against the gold-standard answer. A prediction was deemed correct if the relative error, (prediction $-$ gold) / gold, was less than 1\%. We used \textbf{accuracy}, which is the proportion of questions correctly answered, as our metric. \revision{Crucially, our evaluation leveraged Numbat for \textbf{dimensional analysis}, allowing us to automate the unit conversion~(e.g., $1000\text{ J} = 1\times10^3\text{ N}\cdot\text{m}$) beyond simple numerical matching.}

\revision{To evaluate the quality of the complete reasoning steps, we also used \textbf{PRM score} to evaluate the reasoning steps generated by LLMs. There were two evaluation settings: \textbf{one-step} and \textbf{multi-step}. One-step evaluation scored the overall output using PRM, whereas multi-step evaluation divided the output into multiple steps, scored each individually, and took the average as the final score. Following the setting of DPO mentioned above, here we again used Qwen2.5-Math-PRM-7B as the PRM. We adopted PRM scoring instead of strict formula matching because valid reasoning paths in formula-based problems are often non-unique; different formula combinations can lead to the correct answer.}

\section{Experimental Results}

We present the evaluation results on the Chinese version of \datasetname. The results on FormulaReasoning+ are in Table~\ref{tab:llms_results_enhanced}. The results on the English version are in Appendix~\ref{appendix:en_exp_results}, which are consistent with those on the Chinese version. The results of PoT are in~\ref{appendix:cmp_cot_pot}, not better than CoT.

\begin{table}[th]
\centering
\small
\caption{Accuracy of LLMs with CoT prompts.}
\begin{adjustbox}{width=0.7\linewidth}
\begin{tabular}{p{2.8cm}rrr}
    \toprule
    Model & HoF & HeF & Avg. \\
    \midrule
    ERNIE-4.5-21B-A3B & 66.34 & 67.96 & 67.12 \\
    Gemma-3n-E4B-it & 14.04 & 14.47 & 14.25 \\
    GPT-oss-20B & 85.96 & 82.95 & 84.50 \\
    GPT-4o & 69.49 & 65.37 & 67.50 \\
    GLM-4-plus & 61.74 & 63.57 & 62.63 \\
    Qwen3-235B-A22B & 80.63 & 83.98 & 82.25 \\
    DeepSeek-R1 & 88.62 & 91.47 & 90.00 \\
    o3-mini & 92.01 & 89.66 & 90.87 \\
    GPT-5 & 92.01 & 92.76 & 92.37 \\
    \midrule
    Human & 93.49 & 90.47 & 92.03 \\
    \bottomrule 
\end{tabular}
\end{adjustbox}
\label{tab:llms_results}
\end{table}

\begin{table}[th]
\centering
\small
\caption{\revision{Accuracy of LLMs with CoT prompts on \datasetname+.}}
\begin{adjustbox}{width=0.88\linewidth}
\begin{tabular}{p{3.5cm}rrr}
    \toprule
    Model & HoF & HeF & Avg. \\
    \midrule
    ERNIE-4.5-21B-A3B & 6.89 & 6.08 & 6.48 \\
    Gemma-3n-E4B-it & 0.99 & 1.32 & 1.16 \\
    GPT-oss-20B & 59.36 & 57.41 & 58.38 \\
    GPT-4o & 15.27 & 17.72 & 16.50 \\
    GLM-4-plus & 17.98 & 16.67 & 17.32 \\
    Qwen3-235B-A22B & 78.08 & 77.78 & 77.93 \\
    DeepSeek-R1 & 83.99 & 80.42 & 82.20 \\
    o3-mini & 73.89 & 73.96 & 73.93 \\
    GPT-5 & 76.11 & 73.28 & 74.69 \\
    \bottomrule
\end{tabular}
\end{adjustbox}
\label{tab:llms_results_enhanced}
\end{table}

\subsection{Results of LLMs with CoT Prompts}
\label{sec:mainllmcot}
We evaluated a diverse range of LLMs, as shown in Table~\ref{tab:llms_results} and Table~\ref{tab:llms_results_enhanced}.
Our analysis reveals that model performance correlates with both parameter scale and architectural design. As shown in Table~\ref{tab:llms_results}, large-scale models (e.g., DeepSeek-R1 and o3-mini) achieved competitive results with accuracies exceeding 88\% on both HoF and HeF metrics. However, substantial performance gaps existed for smaller-scale models, where architectures like Gemma-3N-E4B-it exhibit accuracies below 15\%.
Notably, despite their impressive capabilities, even advanced large-scale models such as GPT-4o~(67.50\%), GLM-4-plus~(62.63\%), and Qwen3-235B-A22B~(82.25\%) still trailed behind human performance (92.03\%).
This significant variance suggests that \emph{while parameter scaling yields notable improvements, there remains considerable optimization potential for resource-efficient models}. Future work should investigate architectural adaptations and training strategies specifically tailored for compact models, which are critical for practical deployment scenarios where computational resources are constrained. The observed performance spectrum underscores the importance of scale-aware evaluation in formula-based reasoning research.

\revision{The results on \datasetname+, as shown in Table~\ref{tab:llms_results_enhanced}, demonstrate the increased difficulty of the extended dataset. All models experienced a drop in performance. Notably, the accuracy of GPT-4o and GLM-4-plus declined sharply to 16.50\% and 17.32\% respectively. Even GPT-5 showed a decrease of 17.68\%, highlighting the challenges posed by the more complex reasoning tasks.}

We provide an error analysis in Appendix~\ref{appendix:error_analysis}.

PRM scores to assess the reasoning steps generated by LLMs are included in Appendix~\ref{appendix:process_reward_model_evaluation}. %

\subsection{Results of LLMs with Formula Retriever}

We found that \emph{incorporating retrieved formulas into the prompts improved the performance of LLMs}. For example, for ERNIE-4.5-21B-A3B, the accuracy increased by~0.72 from 66.34 to 67.06 on the HoF. Note that it would be meaningless to carry out this experiment on the HeF test set where the required formulas were not included when training the retriever.

Our investigation found that the top five formulas retrieved often included irrelevant ones, as the number of formulas required varied for different questions. The presence of extraneous formulas might introduce noise, suggesting considerable room for improvement in structured formula retrieval.

\begin{table}[ht]
\small
\centering
\caption{\revision{Accuracy of supervised fine-tuned models.}}
\begin{adjustbox}{width=\linewidth}
\begin{tabular}{llll}
    \toprule
    Model & HoF & HeF & Avg. \\
    \midrule
    Qwen2.5-Math-7B-Instruct & 65.62 & 58.91 & 62.27 \\
    \quad + CoT-SFT & 72.64~\red{\scriptsize $+7.02$} & 65.63~\red{\scriptsize $+6.72$} & 69.14~\red{\scriptsize $+6.87$} \\
    \quad + DA & 67.55~\red{\scriptsize $+1.93$} & 62.53~\red{\scriptsize $+3.62$} & 65.04~\red{\scriptsize $+2.77$} \\
    \midrule
    Qwen2.5-Math-1.5B-Instruct  & 64.41 & 55.56 & 59.99 \\
    \quad + CoT-SFT & 66.34~\red{\scriptsize $+1.93$} & 53.75~\green{\scriptsize $-1.81$} & 60.05~\red{\scriptsize $+0.06$} \\
    \quad + DA & 66.59~\red{\scriptsize $+2.18$} & 59.17~\red{\scriptsize $+3.61$} & 62.88~\red{\scriptsize $+2.89$} \\
    \bottomrule 
\end{tabular}
\end{adjustbox}
\label{tab:small_models_results}
\end{table}
\subsection{Results of Supervised Fine-Tuned Models}

Table~\ref{tab:small_models_results} shows the results for the supervised fine-tuned models, with and without CoT-SFT or DA. 
In most settings, both models achieved higher scores on the HoF test set than on HeF, yet they still exhibited considerable performance on the latter. This indicated that the unseen formulas influenced the performance of the models, though they still demonstrated a level of generalizability.

In particular, Qwen2.5-Math-7B-Instruct with CoT-SFT outperformed GPT-4o on \datasetname, \emph{demonstrating the effectiveness of our fine-tuning method: delegating numerical calculations to a calculator and focusing on CoT reasoning}, although such a comparison might not be entirely fair.
Data augmentation also improved the reasoning capacity of small models.

\subsection{Results of DPO (including Results on Other Numerical Reasoning Datasets)}

As shown in Appendix~\ref{appendix:dpo_results}, DPO with the preference data derived from the training set of \datasetname\ generally improved the accuracy of 1.5B--7B models on both \datasetname\ and other numerical reasoning datasets: GSM8K~\citep{gsm8k}, MATH~\cite{hendrycks2021measuring}, SVAMP~\cite{patel-etal-2021-nlp}, and GaoKao2023-en~\cite{chenadvancing}. These models exhibited much lower accuracy on \datasetname\ than those datasets. \emph{The results underscore the distinct characteristics of our dataset compared to other numerical reasoning benchmarks and its additional value as a potential DPO resource.}\\

\section{Conclusion}
\label{sec:conclusion}

We introduced \datasetname, a bilingual dataset for formula-based numerical reasoning. For each question, we annotated its complex reasoning steps with normalized physics formulas. We constructed a consolidated formula database after merging equivalent formulas, serving as an external knowledge base to be used in RAG.
We used our dataset to evaluate LLMs of various sizes, RAG, fine-tuned small models, and DPO methods, revealing significant performance variations across model scales.
Our findings highlight that while larger models show promising performance, smaller-scale LLMs still face significant challenges in formula-based numerical reasoning, indicating opportunities for further advancements in multi-step reasoning guided by domain knowledge.

Future work will use \datasetname's formula knowledge to enhance LLM numerical reasoning via knowledge-driven or reinforcement learning methods. Moreover, given the domain-agnostic nature of our construction framework, we aim to extend our implementation to other scientific and engineering domains, such as chemistry and finance, to build a more comprehensive multi-disciplinary formula reasoning benchmark. %

\newpage
\section*{Limitations}
One limitation of our work is that although \datasetname\ provides step-level supervision information~(i.e., formulas), due to the diversity of reasoning paths and formula expressions, we only used PRM to perform step-level process evaluation. Another limitation is that our dataset is focused on physics. We chose junior high school physics because it is not too hard to be understood by ordinary people, which benefited our annotation and evaluation efforts. It is possible to explore formula-based numerical reasoning in other domains such as chemistry and engineering.

\bibliography{custom,anthology}

\appendix
\clearpage
\section{Appendix}
\label{sec:appendix}

\subsection{Related Work}
\label{sec:related_work}

\subsubsection{Numerical Reasoning Datasets}
The study of numerical reasoning in natural language has existed for years. Numerous datasets, such as GSM8K~\citep{cobbe2021training}, TSQA~\citep{li2021tsqa}, and MATH~\citep{hendrycks2021measuring}, have addressed natural language numerical reasoning. 
Another line of research that focuses on numerical reasoning in natural language is math word problem~(MWP). MWP tasks typically provide a short passage~(i.e., a question) and require the generation of an arithmetic expression that can compute an answer. Representative datasets include MAWPS~\citep{koncel-kedziorski-etal-2016-mawps}, Math23K~\citep{wang-etal-2017-deep}, MathQA~\citep{amini-etal-2019-mathqa}, etc.
Several works focus on numerical reasoning in specialized domains. Examples include GeoSQA~\citep{geosqa}, which focuses on the geography domain, STEM~\citep{stem} and ScienceQA~\citep{scienceqa} which cover multiple disciplines in science and technology. \emph{Distinguished from these datasets, the numerical reasoning questions in our \datasetname\ have explicitly labeled formulas}.

The recently introduced datasets Math23K-F and MAWPS-F~\citep{Liu2023GuidingMR} require formulas for 33\% and 38\% of the questions, respectively, but their formulas are all simple commonsense formulas~(e.g., total\_cost = unit\_cost $\times$ total\_number). \emph{By contrast, our \datasetname\ adapts questions from physics exams, with every question accompanied by a fine-grained annotation of domain-specific formulas. In addition, we provide a consolidated formula database that can serve as an external knowledge base to assess RAG solutions.}

\subsubsection{Numerical Reasoning Methods}
Methods for numerical reasoning have evolved from statistical approaches~\citep{hosseini-etal-2014-learning,kushman-etal-2014-learning} to those based on rules and templates~\citep{shi-etal-2015-automatically,wang2019template} and further to deep learning models~\citep{gupta2019neural,chen-etal-2022-teaching-neural,kim-etal-2022-exploiting,li2023dyrren}. In recent years, with the rapid development of LLMs, they have demonstrated strong capabilities to resolve numerical reasoning questions. Consequently, several methods have been proposed to improve the reasoning abilities of LLMs, including the notable CoT method~\citep{cot}, along with many subsequent variant approaches~\citep{cot_zero_shot,wang2022self,zhou2022least,li-etal-2023-making}. Preference learning methods have also emerged~\citep{NEURIPS2023_a85b405e}.

Our experiments implemented and evaluated representative existing methods as baselines for \datasetname, including zero-shot CoT prompting for LLMs ranging from 4B to more than 100B parameters. We trained a specialized formula retriever to be used in RAG. We divided the reasoning process into formula generation, parameter extraction, and numerical calculation, and used data augmentation to enhance fine-tuned small models. We derived preference data and performed DPO.

\subsection{Appendix for Dataset Construction}
\subsubsection{Prompts for Formula Normalization}
\label{appendix:prompts_formula_standardization}

\begin{figure*}[!h]
    \centering
    \includegraphics[width=0.9\linewidth]{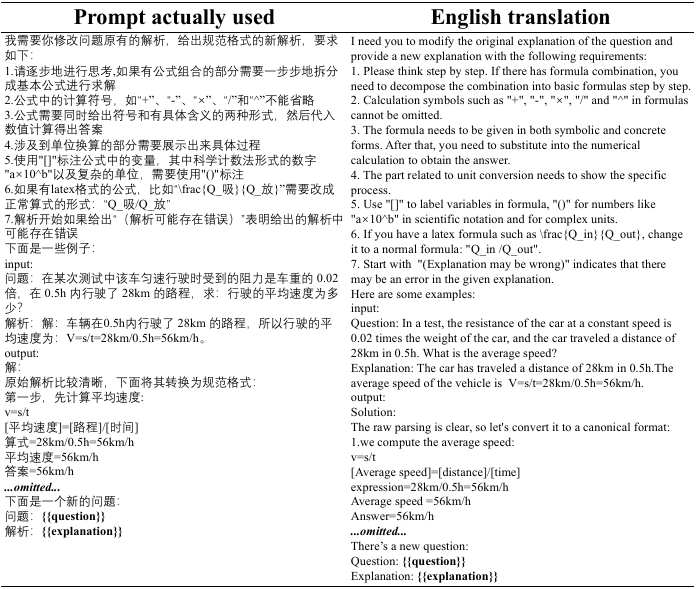}
    \caption{Prompt for explanation normalization.}
    \label{fig:LLMprompt1}
\end{figure*}

\begin{figure*}
    \centering
    \includegraphics[width=0.9\linewidth]{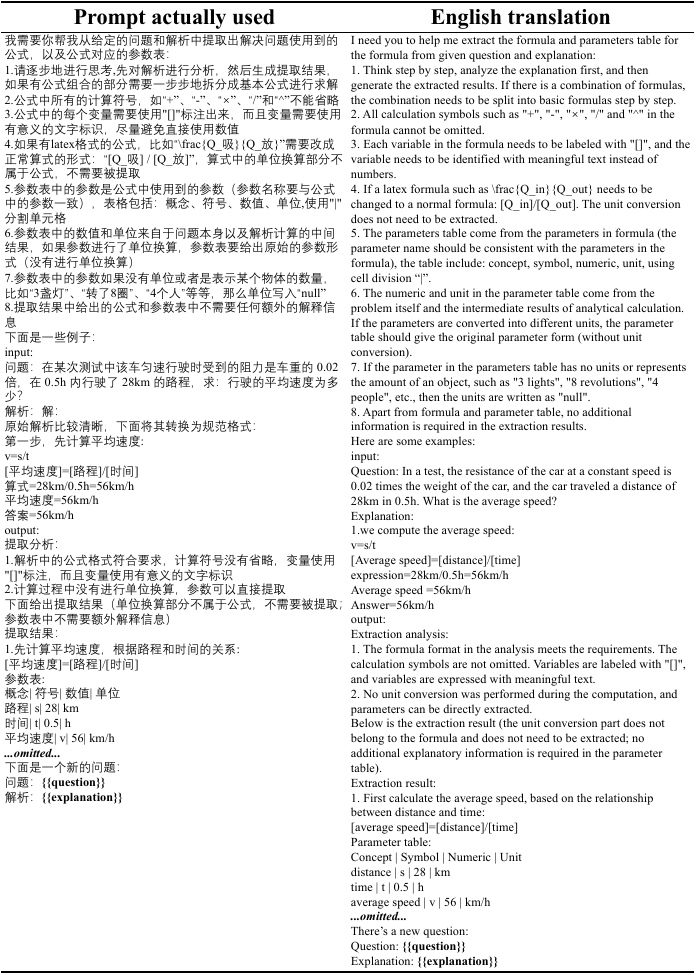}
    \caption{Prompt for parameter extraction.}
    \label{fig:LLMprompt2}
\end{figure*}

\begin{figure*}
    \centering
    \includegraphics[width=0.9\linewidth]{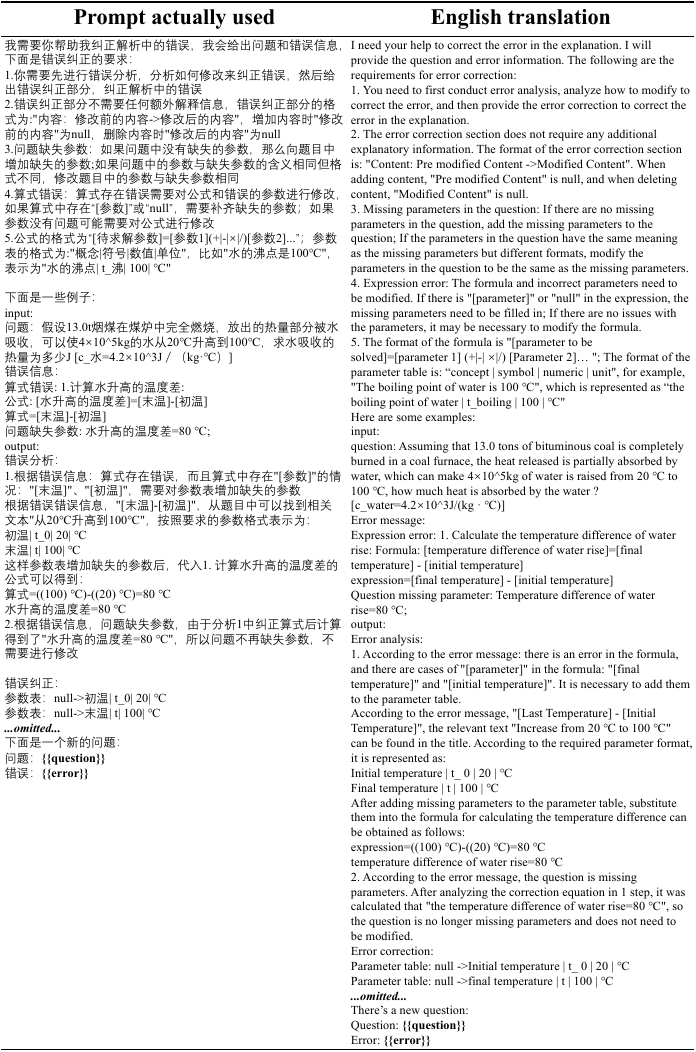}
    \caption{Prompt for correcting calculation errors.}
    \label{fig:LLMprompt3}
\end{figure*}

The formula normalization process followed a coarse-to-fine annotation process as described in Section~\ref{sec:formula_normalization}. In the coarse-grained annotation phase, the natural language explanations were normalized and the associated parameters were extracted from these explanations. These prompts are provided in Figures~\ref{fig:LLMprompt1} and \ref{fig:LLMprompt2}. The fine-grained annotation phase focused on error correction, which was divided into three specific categories: input errors, where the parameters mentioned in the explanation were absent from the question; calculation errors, which occurred when the Numbat calculator reported an error during the computation process; and output errors, where the final calculated answer was incorrect. We provide a prompt for correcting the calculation errors as an example in Figure~\ref{fig:LLMprompt3}; the prompts for the other two error types are provided in our GitHub repository. The entire normalization procedure used a few-shot prompt with six examples.

\subsubsection{Examples of Removed Questions}
\label{appendix:examples_of_deleted_questions}
\begin{figure*}[!h]
    \centering
    \includegraphics[width=0.8\linewidth]{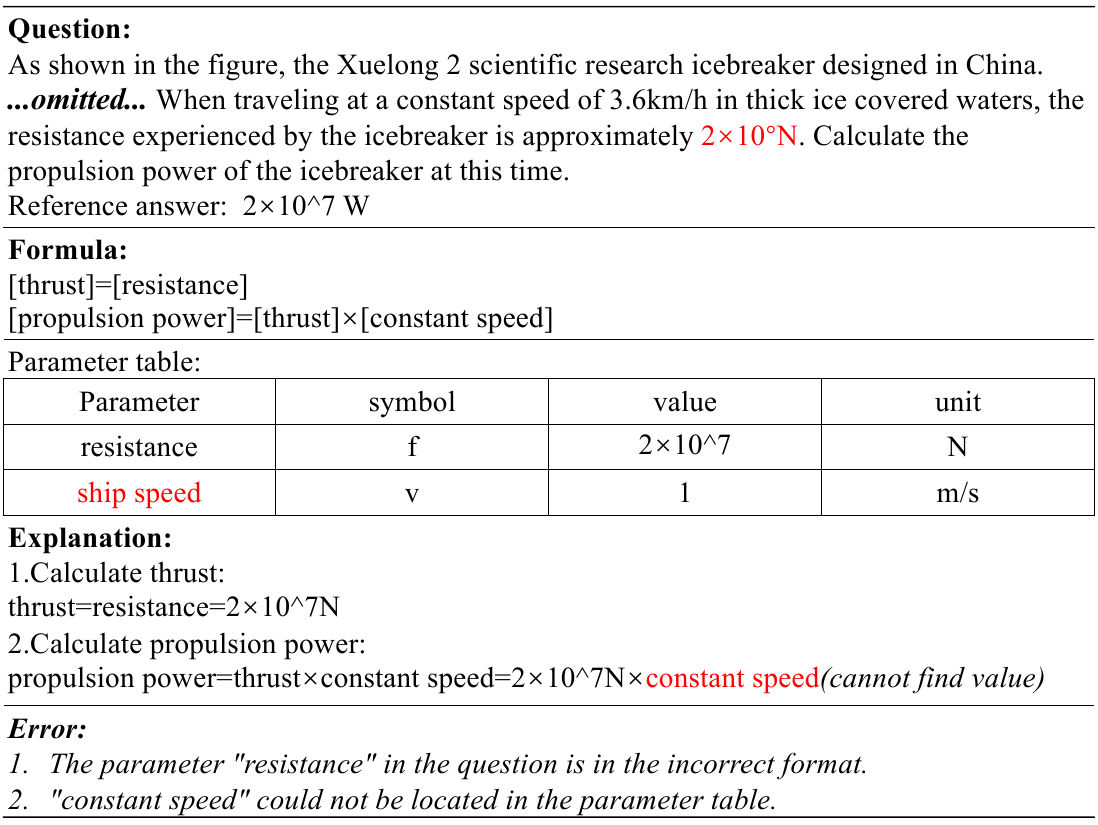}
    \caption{An example of removed question.}
    \label{fig:example_deleted_question}
\end{figure*}

The questions that remained incorrect despite multiple attempts by the LLM were of notably poor quality, e.g. missing important reasoning steps or having an incorrect reference answer. We provide an example of these questions in Figure~\ref{fig:example_deleted_question}.

\subsubsection{Prompt for Semantics Based Merging}
\label{appendix:semantic_based_merging}
\begin{figure*}[!h]
    \centering
    \includegraphics[width=0.8\linewidth]{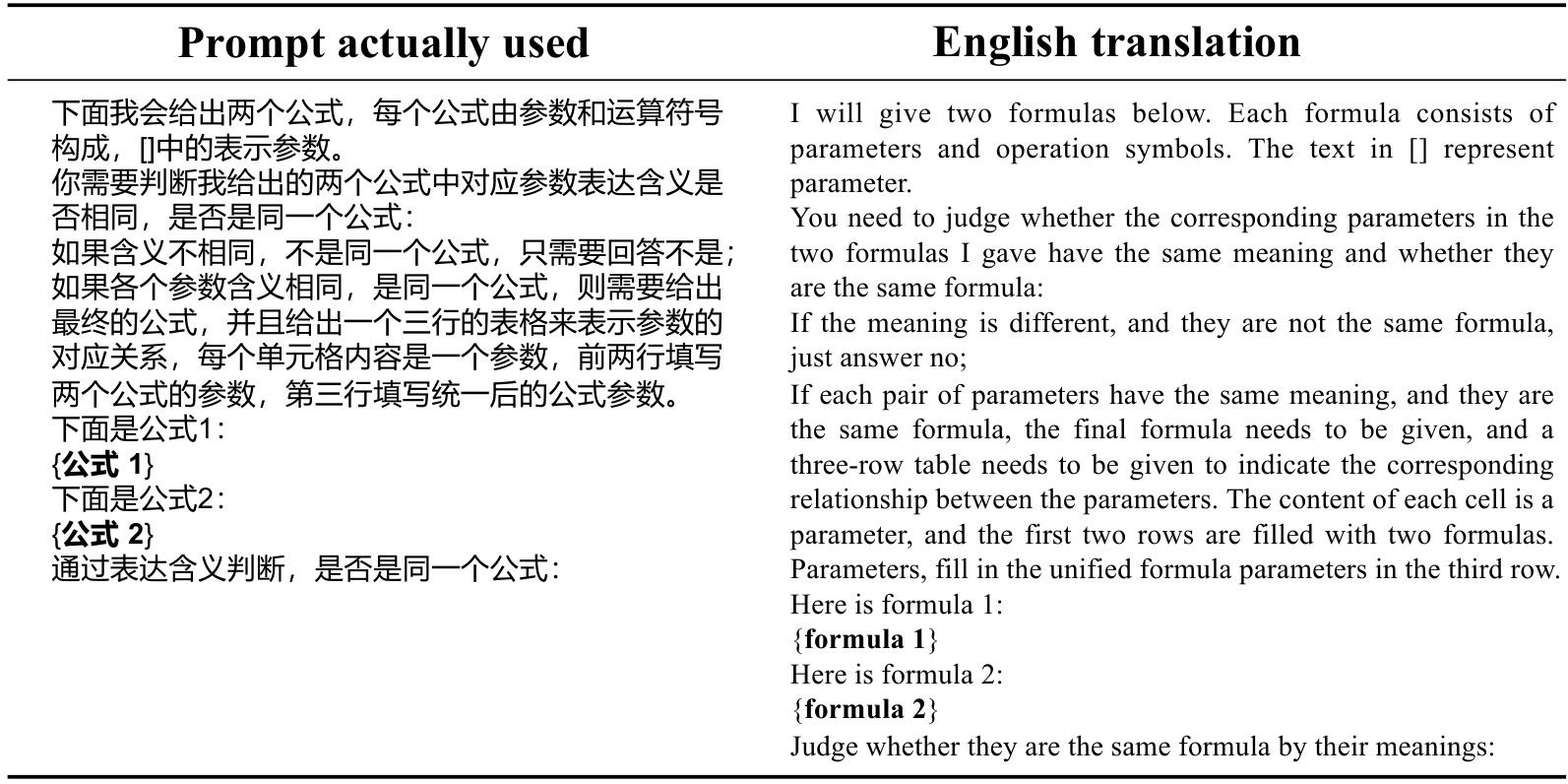}
    \caption{Prompt for semantics based merging.}
    \label{fig:semantic_based_merging_prompt}
\end{figure*}

Semantics based merging primarily employed the LLM to comprehend formulas, determined if two formulas were semantically equivalent, and subsequently determined whether they could be merged into a single formula. The prompt for this procedure is illustrated in Figure~\ref{fig:semantic_based_merging_prompt}. This approach ensured that nuanced meanings embedded within formulas were accurately captured and evaluated for potential merging, thereby enhancing the quality of the formula database.

\subsubsection{Prompt for Dataset Translation}
\label{appendix:dataset_translation_prompt}
\begin{figure*}[!h]
  \centering
  \begin{subfigure}{\linewidth}
  \centering
    \includegraphics[width=0.8\linewidth]{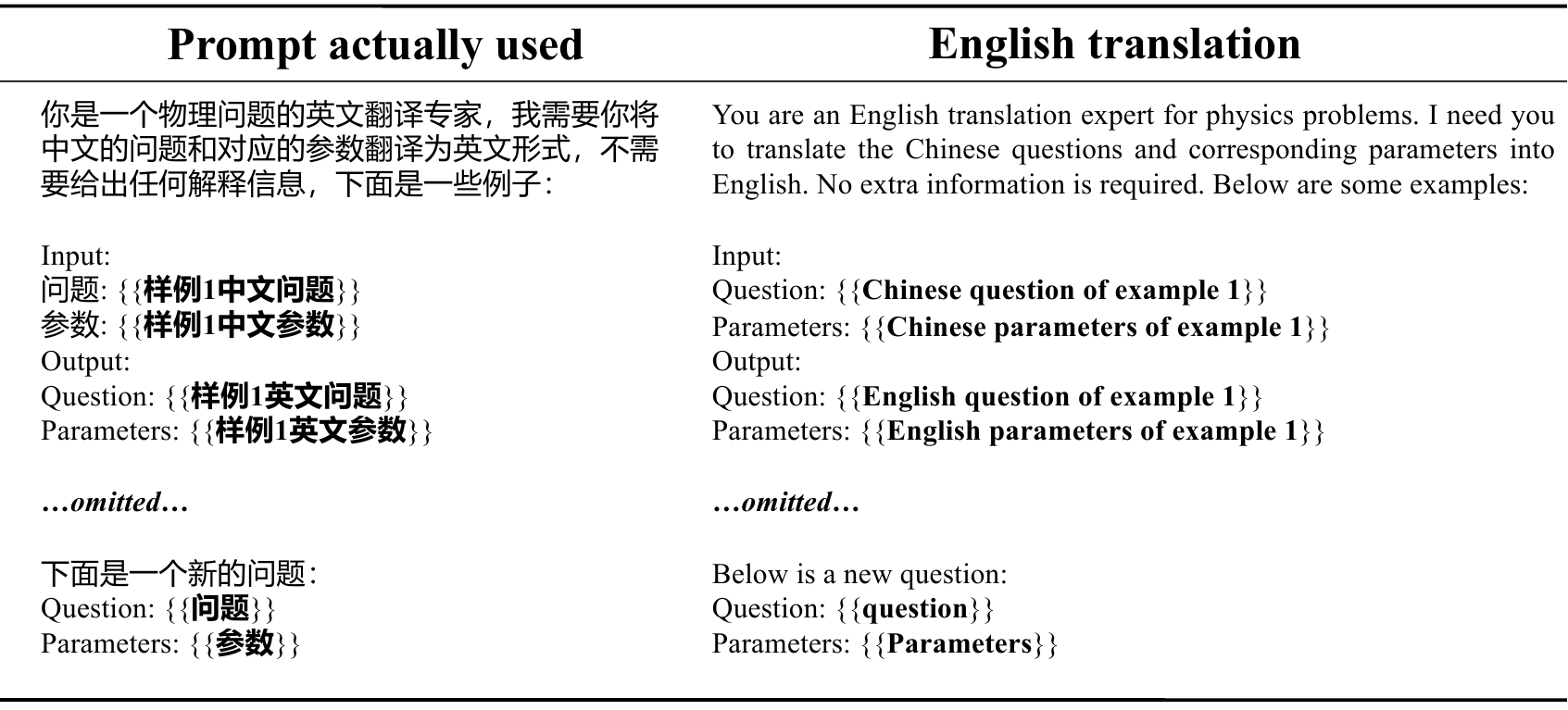}
  \end{subfigure}
  \hfill
  \begin{subfigure}{\linewidth}
  \centering
    \includegraphics[width=0.8\linewidth]{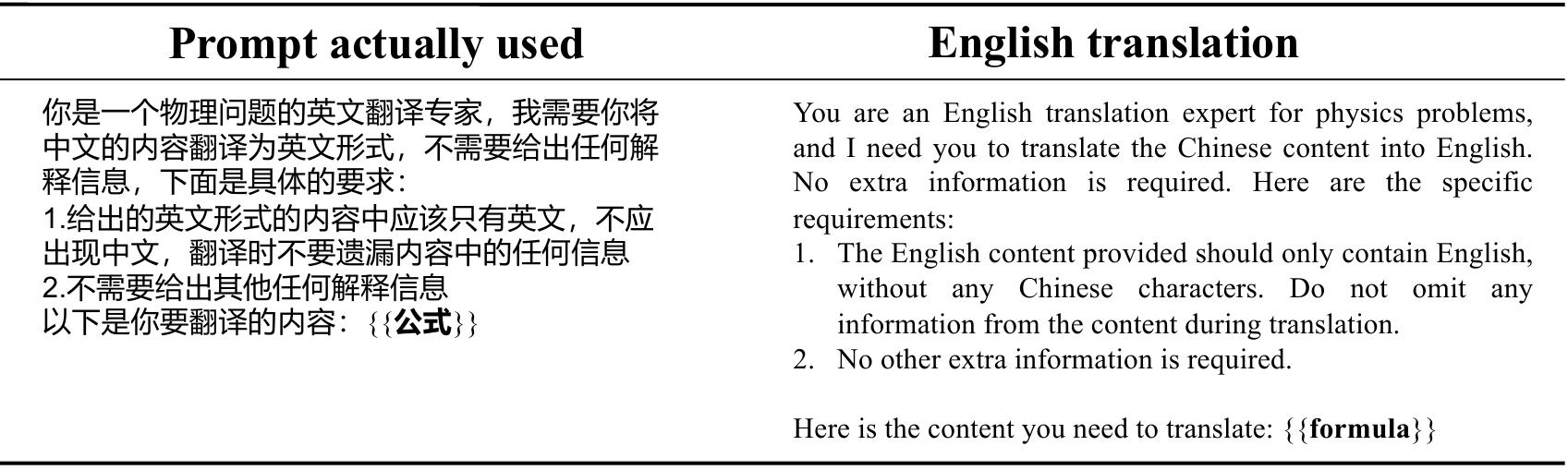}
  \end{subfigure}
  \caption{Prompts for translating questions (top) and formulas (bottom).}
  \label{fig:translation_prompt}
\end{figure*}

The prompts for translating the dataset are provided in Figure~\ref{fig:translation_prompt}.

\subsubsection{\revision{Question Coupling}}
\label{appendix:question_coupling}

In this section, we provide a detailed walkthrough of the \emph{Question Coupling} method using a concrete example. This method increased reasoning difficulty by coupling the original question with a generated auxiliary question through a mathematical expression. All the questions had been verified by Numbat to meet the constraints and be solvable.

\paragraph{Original Question}
Consider the following original question from \datasetname:
\begin{quote}
\textbf{Question:} A block of aluminum with a mass of 50 g releases 880 J of heat, and its temperature decreases to 12$^\circ$C. What was the original temperature of the aluminum block? [$C_{aluminum} = 0.88\times 10^3$ J/(kg$\cdot^\circ$C)]\\
\textbf{Answer:} 32$^\circ$C
\end{quote}

\paragraph{Step 1: Parameter Masking}
We first identified mutable parameters in the question. In this example, we randomly selected the heat released (880 J) as the parameter to mask, denoted as $[X]$. We also designated the original answer (32$^\circ$C) as the second unknown parameter $[Y]$.
\begin{itemize}
    \item $[X]$: Heat released (880)
    \item $[Y]$: Original temperature (32)
\end{itemize}

\paragraph{Step 2: Expression Construction}
We constructed a linear expression involving $[X]$ and $[Y]$. To ensure numerical stability and reasonable magnitude, we adjusted the coefficients based on the order of magnitude of the parameters.
In this example, we aligned the magnitudes and constructed the expression:
\[ E(X, Y) = 50 \cdot Y - 1 \cdot X \]
Substituting the values:
\[ 50 \times 32 - 1 \times 880 = 1600 - 880 = 720 \]
The result of the expression is 720.

\paragraph{Step 3: Auxiliary Question Generation}
We then generated an independent physics question whose answer corresponds to the calculated result (720). We randomly assigned a unit (e.g., Newtons) to this value.
Using GPT-4o, we generated the following auxiliary question:
\begin{quote}
\textbf{Auxiliary Question:} Ming is conducting a physics experiment. He needs to use a crane to lift a mass of 80 kg vertically upwards at a constant speed. Given that the local gravitational acceleration is 9.0 m/s$^2$, what force does the crane need to exert to make the mass rise at a constant speed?
\end{quote}
The answer to this auxiliary question is $F = mg = 80 \times 9.0 = 720$ N, which matches our target value.

\paragraph{Step 4: Final Compilation}
Finally, we assembled a new hybrid question. We injected a distractor parameter (e.g., ``calorific value of gas'') to further increase difficulty. The final question requires the model to solve the auxiliary question to find the value of the expression~(i.e.,~720 here), and then solve a system of equations in $X$ and $Y$ where one equation represents their relation in the original question and the other is, in this example, $720 = E(X, Y)$.

\begin{quote}
\textbf{Final Question:} A block of aluminum with a mass of 50 g releases $[X]$ amount of heat and its temperature decreases to 12$^\circ$C. What was the original temperature of the aluminum block? [$C_{aluminum} = 880$ J/(kg$\cdot^\circ$C)] The answer to the above question is $[Y]$.\qquad
Under the International System of Units (with the exception that temperature is expressed in degrees Celsius), the value of the numerical expression $50Y - X$ of the physical quantities corresponding to $[X]$ and $[Y]$, is the same as the result of the following independent question: Ming is conducting a physics experiment. He needs to use a crane to lift a mass of 80 kg vertically upwards at a constant speed. Given that the local gravitational acceleration is 9.0 m/s$^2$, what force does the crane need to exert to make the mass rise at a constant speed?
\end{quote}

\subsection{Appendix for Experimental Setup}

\subsubsection{Evaluated LLMs}
\label{appendix:evaluated_llms}
We conducted experiments across a diverse spectrum of models including cost friendly small-scale models including ERNIE-4.5-21B-A3B~\cite{ernie45}, Gemma-3n-E4B-it~\cite{gemma_3n_2025}, GPT-oss-20B~\citep{gptoss}, alongside high-cost LLMs such as GPT-4o~\citep{gpt4o}, GLM-4-plus~\citep{glm4}, Qwen3-235B-A22B~(Qwen3-235B-A22B-Thinking-2507)~\citep{qwen3}, DeepSeek-R1~\citep{deepseekai2025deepseekr1incentivizingreasoningcapability}, o3-mini~\citep{o3mini} and GPT-5~\citep{openai2025gpt5}.

\subsubsection{Prompt for Evaluated LLMs}
\label{appendix:zero_few_shot_prompt_example}

\begin{figure*}[!h]
  \centering
    \includegraphics[width=0.8\linewidth]{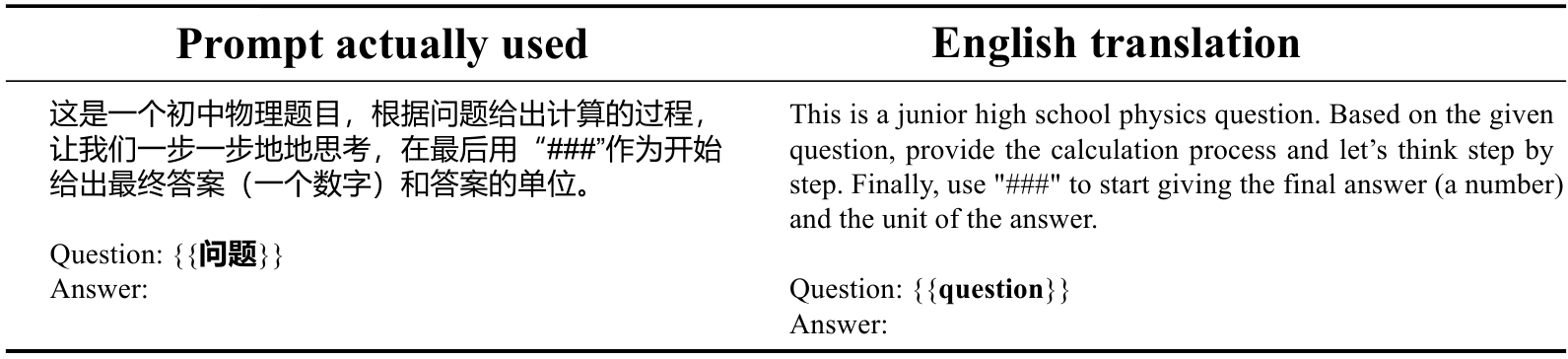}
  \caption{Prompt for evaluated LLMs.}
  \label{fig:zero_few_shot_prompts}
\end{figure*}

The prompt is provided in Figure~\ref{fig:zero_few_shot_prompts}.

\subsubsection{Formula Retriever}
\label{appendix:formula_retriever}

\paragraph{Implementation}
Let the number of formulas in the formula database be $N$. During training, we randomly initialized a matrix $\mathbf{F}\in\mathbb{R}^{N\times d}$, where $d$ is the hidden size and the $i$-th row in $\mathbf{F}$ represented the initial representation of the $i$-th formula in formula database. We denoted a batch of questions with a batch size of $B$ as $Q=\{q_1, q_2, ..., q_B\}$. The indices of the gold-standard formulas corresponding to these $B$ questions were denoted as $L=\{l_1,l_2,\cdots,l_B\}$~(i.e. the label of $q_i$ is $l_i$, where $1\le i\le B$).

BERT was utilized to encode each question,
\begin{equation}
    \mathbf{h}^i_{cls},  \mathbf{h}^i_{1}, \cdots = \mathtt{BERT}(q_i), 1\le i\le B.
\end{equation}
Subsequently, we took the CLS vector $\mathbf{h}^i_{cls}$ as the representation for the $i$-th question.

We utilized in-batch negatives and contrastive learning loss,
\begin{equation}
    \mathcal{L}=-\frac{1}{B}\sum_{1\le i\le B}\log\frac{\exp(\cos(\mathbf{h}^i_{cls}, \mathbf{F}_{l_i}))}{\sum_{1\le j\le B}\exp(\cos(\mathbf{h}^i_{cls}, \mathbf{F}_{l_j}))}.
\end{equation}
Each question might correspond to multiple correct formulas, and we ensured that the same question did not appear twice in the same batch when loading the data. Based on the implementation of Chinese-BERT-wwm-base, we tested the retrieval performance on the HoF test set and found that Recall@5 reached 82.46\%.

\begin{figure*}[!h]
  \centering
    \includegraphics[width=0.8\linewidth]{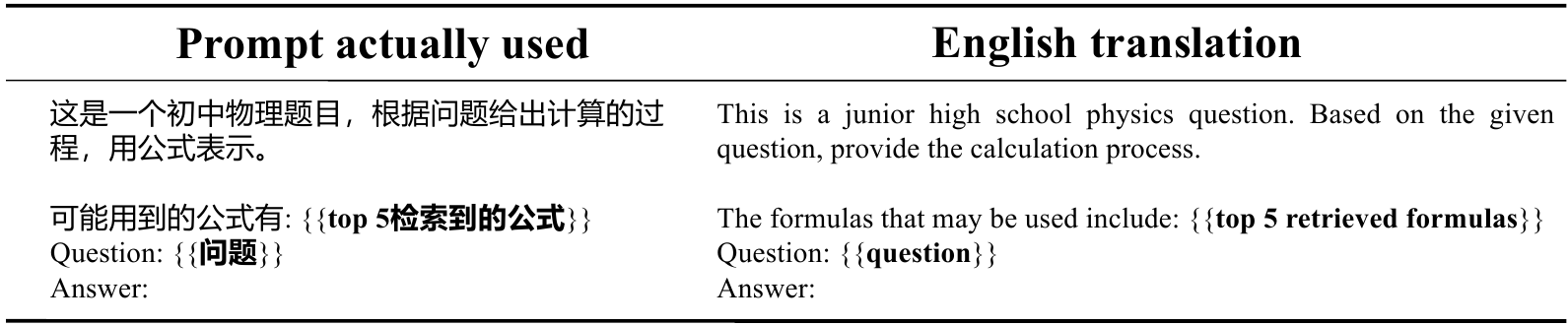}
  \caption{Prompt for evaluated LLMs with formula retriever.}
  \label{fig:zero_few_shot_prompts_FR}
\end{figure*}

\paragraph{Prompt for Evaluated LLMs with Formula Retriever}
\label{appendix:few_shot_prompts_for_llms_with_formula_retriever}
For each question, we included the top-5 retrieved formulas in the prompt provided in Figure~\ref{fig:zero_few_shot_prompts_FR}.

\subsubsection{Data Augmentation for Supervised Fine-Tuned Models} 
\label{appendix:data_augmentation}

Several studies~\citep{2024arXiv240302990D,zheng-etal-2023-augesc,whitehouse2023llmpowered} use LLM for data augmentation, which mainly focus on daily conversations or sentiment analysis and do not require rigorous numerical calculations. Some research~\citep{shum-etal-2023-automatic} on data augmentation that involves numerical calculations employs LLM to generate solutions for existing questions to aid training, rather than to generate new questions. In contrast to these approaches, we generated complete questions that involve numerical calculations~(particularly focusing on formulas), along with automatic improvement and selection to ensure data quality.

We divided the data generation process into the following steps. 
First, we randomly generated 17,000 prompts. Each prompt was obtained by stacking five question-explanation pairs sampled from the training set. At the end of the prompt, the LLM was required to generate the sixth question-explanation pair. 
Second, we normalized the formulas generated. Except for the absence of manual review, the remaining steps were consistent with those in Section~\ref{sec:formula_normalization}. Finally, we utilized Numbat to check whether the calculation process in the data generated by the LLM was correct, and discarded the data with incorrect calculations.
After the above steps, we finally retained about 2,500 questions.

We found that mixing the newly generated data into the original training set did not always bring positive improvement, perhaps because the newly generated data did not undergo manual review. We found that randomly selecting a small portion of the newly generated data could improve the performance of the model.
We set the mixing ratio to 10\% which is an experiential proportion. We fine-tuned each model using the augmented dataset. After training for a fixed number of epochs~(50 epochs), we selected the final checkpoints as DA models.

\subsubsection{Preference Data Generation}
\label{DPO_construction}

We applied the Monte Carlo Tree Search (MCTS) method to the unaltered questions in the training set of \datasetname\ to generate preference data for DPO. It used the reasoning steps as nodes, and used the generation model to expand nodes. Each node had one more reasoning step than its parent node. The root node had no reasoning step, only containing the original question. The termination state was reached when the termination token was generated or the current question was solved. We used the PRM to guide the generation. The PRM score was used as a reward to update the nodes.

After generation, a tree was created for each sample. Each node contained the reasoning steps generated up to that node and the correctness score of the generated steps. We identified the best and worst paths in the tree. Each path was scored by the average score from the root node to the terminated leaf node. Limited by the number of iterations, some leaf nodes might not represent termination and their corresponding paths were not used.

To consider the inclusion of formula knowledge, we further screened the generated data. For each sample, we examined its best and worst paths. If there was no significant formula knowledge on these paths, we would discard this sample. We used a rule-based pattern matching method to identify formula knowledge. This pattern was consistent with the example given in the generation prompt used in MCTS, which was an expression consisting of "[\{parameter\}]".

\subsubsection{Implementation Details}
\label{appendix:implementation_details}

We accessed DeepSeek-R1 through the DeepSeek API\footnote{https://api-docs.deepseek.com/}, Qwen series models through the Alibaba Bailian platform\footnote{https://bailian.console.aliyun.com/}, GLM-4-plus through ZHIPU--AI API\footnote{https://open.bigmodel.cn/dev/api/normal-model/glm-4}, and other models through OpenRouter\footnote{https://openrouter.ai/} with their default hyper-parameters.
We set temperature=0 for direct answer generation and set the maximum output length to~1,024.
In the case of CoT-SFT, which directly outputted formulas along with numerical values and units of parameters, we set temperature=0 to obtain the response with the highest probability and set the maximum output length to~512 to obtain formulas and parameters.
Training Qwen2.5-Math-7B-Instruct(lora) and Qwen2.5-Math-1.5B-Instruct(full) in CoT-SFT used~3 and 8~hours, respectively.

\subsection{Appendix for Experimental Results}

\subsubsection{\revision{English Version of \datasetname}}
\label{appendix:en_exp_results}

The evaluation results for LLMs with CoT prompts on the English \datasetname\ dataset are presented in Table~\ref{tab:llms_results_en}.
Large-scale models (e.g., GPT-5 and DeepSeek-R1) achieved competitive results with accuracies exceeding 87\% on both HoF and HeF metrics. However, substantial performance gaps existed for smaller-scale models, where architectures like Gemma-3n-E4B-it exhibited poor accuracies. We found that Gemma-3n-E4B-it and Qwen3-235B-A22B showed performance drops compared to their results on the Chinese version due to instruction-following issues in the English version, where these models failed to output required units as specified in the prompt.

Notably, despite their impressive capabilities, even the best model GPT-5~(88.63\%) still trailed behind human performance (92.03\% as reported in the main text), falling short by 3.4 percentage points. Other advanced models such as GPT-4o~(75.38\%) and GLM-4-plus~(72.37\%) showed even larger gaps with human performance.

There was a general trend of performance decline on \datasetname+.
Notable drops were seen in models like GPT-4o (from 75.38\% to 13.48\%), o3-mini (from 83.13\% to 68.38\%), and GPT-5 (from 88.63\% to 70.51\%).
This confirmed that the the extended \datasetname+ successfully introduced greater complexity, providing a more rigorous benchmark for evaluating the reasoning capabilities of LLMs.

This substantial performance disparity indicates that although scaling parameters leads to marked improvements, there is still significant room for optimizing resource-efficient models. Future research should prioritize architectural and training optimizations for compact models, which are essential for resource-constrained environments. The performance variation observed across models highlights the critical need for scale-aware evaluation in formula-based reasoning.

\begin{table}
\centering
\small
\caption{Accuracy of LLMs with CoT prompts on English \datasetname.}
\begin{adjustbox}{width=\linewidth}
\begin{tabular}{p{2.8cm}rrrrrr}
    \toprule
    & \multicolumn{3}{c}{\datasetname} & \multicolumn{3}{c}{\datasetname+} \\
    \cmidrule(lr){2-4} \cmidrule(lr){5-7}
    Model & HoF & HeF & Avg. & HoF & HeF & Avg. \\
    \midrule
    ERNIE-4.5-21B-A3B & 63.20 & 56.85 & 60.13 & 7.39 & 9.26 & 8.33 \\
    Gemma-3n-E4B-it & 2.66 & 2.84 & 2.75 & 2.71 & 2.65 & 2.68 \\
    GPT-oss-20B & 86.20 & 84.75 & 85.48 & 53.45 & 57.67 & 55.56 \\
    GPT-4o & 75.79 & 74.94 & 75.38 & 14.53 & 12.43 & 13.48 \\
    GLM-4-plus & 75.79 & 68.73 & 72.37 & 17.73 & 20.11 & 18.92 \\
    Qwen3-235B-A22B & 62.47 & 66.67 & 64.50 & 77.34 & 77.51 & 77.43 \\
    DeepSeek-R1 & 87.17 & 87.60 & 87.38 & 76.85 & 79.37 & 78.11 \\
    o3-mini & 84.02 & 82.17 & 83.13 & 68.23 & 68.52 & 68.38 \\
    GPT-5 & 88.38 & 88.89 & 88.63 & 71.43 & 69.58 & 70.51 \\
    \bottomrule 
\end{tabular}
\end{adjustbox}
\label{tab:llms_results_en}
\end{table}

\subsubsection{PoT Prompts}
\label{appendix:cmp_cot_pot}

The results of GPT-4o using CoT compared to PoT are shown in Table~\ref{tab:llms_results_pot}. CoT consistently outperformed PoT. One possible reason was that the effectiveness of PoT could be influenced by a hybrid of the language and coding capabilities of LLMs, but that of CoT was only related to the language capabilities. Moreover, for LLMs with strong calculation capabilities, the advantage of using Python to perform numerical calculations might be eroded. \revision{The observed underperformance of PoT compared to CoT appears to also stem from the models' faced challenge in accurate application of formulas within a programmatic context. Therefore, on \datasetname, the bottleneck for current LLMs was mainly in their ability to apply formulas, rather than in calculation.}

\begin{table}[!h]
\small
\centering
\caption{Accuracy of LLMs with CoT and PoT prompts.}
\begin{tabular}{lrrr}
    \toprule
    Model &  HoF test & HeF test & Avg. \\
    \midrule
    GPT-4o~(CoT)  & 69.49 & 65.37 & 67.50 \\
    GPT-4o~(PoT) & 54.72 & 41.34 & 48.25 \\
    \bottomrule
\end{tabular}
\label{tab:llms_results_pot}
\end{table}

\subsubsection{PRM Scores}
\label{appendix:process_reward_model_evaluation}

Tables~\ref{tab:one_step_eval_results_of_llms} and~\ref{tab:multi_step_eval_results_of_llms} show the PRM scores in one-step and multi-step settings, respectively.

\begin{table}[!h]
\small
\centering
\caption{PRM score (one-step) of LLMs with CoT prompts.}
\begin{adjustbox}{width=\linewidth}
\begin{tabular}{lrrr}
    \toprule
    Model &  HoF test & HeF test & Avg. \\
    \midrule
    ERNIE-4.5-21B-A3B & 0.7753 & 0.7582 & 0.7668 \\
    Gemma-3n-E4B-it & 0.3695 & 0.3660 & 0.3678 \\
    GPT-oss-20B & 0.5704 & 0.5203 & 0.5454 \\
    GPT-4o & 0.6150 & 0.5583 & 0.5867 \\
    GLM-4-plus & 0.6728 & 0.6591 & 0.6660 \\
    Qwen3-235B-A22B & 0.6502 & 0.6476 & 0.6489\\
    DeepSeek-R1 & 0.7395 & 0.7393 & 0.7394 \\
    o3-mini & 0.7537 & 0.7556 & 0.7547 \\
    GPT-5 & 0.8196 & 0.8249 & 0.8223 \\
    \bottomrule 
\end{tabular}
\end{adjustbox}
\label{tab:one_step_eval_results_of_llms}
\end{table}

\begin{table}[!h]
\small
\centering
\caption{PRM score (multi-step) of LLMs with CoT prompts.}
\begin{adjustbox}{width=\linewidth}
\begin{tabular}{lrrr}
    \toprule
    Model &  HoF test & HeF test & Avg. \\
    \midrule
    ERNIE-4.5-21B-A3B & 0.8695 & 0.8711 & 0.8703 \\
    Gemma-3n-E4B-it & 0.7366 & 0.7360 & 0.7363 \\
    GPT-oss-20B & 0.5738 & 0.5266 & 0.5502 \\
    GPT-4o & 0.8655 & 0.8444 & 0.8550 \\
    GLM-4-plus & 0.8726 & 0.8738 & 0.8732 \\
    Qwen3-235B-A22B & 0.7946 & 0.7996 & 0.7971\\
    DeepSeek-R1 & 0.7603 & 0.7535 & 0.7569 \\
    o3-mini & 0.8104 & 0.8032 & 0.8068 \\
    GPT-5 & 0.8196 & 0.8248 & 0.8222 \\
    \bottomrule 
\end{tabular}
\end{adjustbox}
\label{tab:multi_step_eval_results_of_llms}
\end{table}

\begin{table*}[ht]
\small
\centering
\caption{Accuracy of DPO on \datasetname\ and other datasets.}
\begin{tabular}{llllll}
    \toprule 
    Model  & GSM8K & MATH & SVAMP & Gaokao2023-en & FormulaReasoning\\
    \midrule
    DeepSeek-R1-Distill-Qwen-7B & 87.7 & 58.8 & 87.7 & 55.3 &34.0 \\
    \quad +DPO & 85.8~\green{\scriptsize $-1.9$} & 59.7~\red{\scriptsize $+0.9$} & 88.1~\red{\scriptsize $+0.4$} &55.8~\red{\scriptsize $+0.5$} &37.0~\red{\scriptsize $+3.0$} \\
    \midrule
    DeepSeek-R1-Distill-Qwen-1.5B & 79.8 & 68.2 & 85.7 & 55.3 & 7.23 \\
    \quad +DPO & 81.3~\red{\scriptsize $+1.5$} & 68.0~\green{\scriptsize $-0.2$} & 86.3~\red{\scriptsize $+0.6$} & 55.1~\green{\scriptsize $-0.2$} &10.47~\red{\scriptsize $+3.24$} \\
    \midrule
    Qwen2.5-Math-1.5B-Instruct & 85.6 & 76.1 & 93.3 & 66.0 & 4.94\\
    \quad +DPO & 85.8~\red{\scriptsize $+0.2$} & 76.5~\red{\scriptsize $+0.4$} & 93.7~\red{\scriptsize $+0.4$} & 65.2~\green{\scriptsize $-0.8$} &7.52~\red{\scriptsize $+2.58$} \\
    \midrule
    Qwen2.5-1.5B-Instruct & 72.9 & 55.0 & 85.5 & 47.5 & 5.94 \\
    \quad +DPO & 73.1~\red{\scriptsize $+0.2$} & 56.2~\red{\scriptsize $+1.2$}& 86.6~\red{\scriptsize $+1.1$}& 47.8~\red{\scriptsize$+0.3$}& 6.57~\red{\scriptsize$+0.63$} \\
    \bottomrule 
    \end{tabular}
\label{tab:dpo_results_on_other_dataset}
\end{table*}
\subsubsection{DPO}
\label{appendix:dpo_results}

We employed the preference data derived from the training set of \datasetname\ to perform DPO with several small models on not only the test sets of \datasetname\ but also other numerical reasoning datasets. Table~\ref{tab:dpo_results_on_other_dataset} shows the results. DPO improved accuracy on \datasetname\ and also improved accuracy in most cases on other datasets, showing an additional value of the preference data derived from our dataset. Notably, although these models exhibited good performance on other datasets, their low accuracy on \datasetname\ indicated that they did not truly master formula-based reasoning required in the physics domain, characterizing \datasetname\ as a valuable dataset presenting unique challenges for LLMs.

\subsection{Error Analysis}
\label{appendix:error_analysis}

\begin{figure*}[!h]
  \centering
  \begin{subfigure}{\linewidth}
    \includegraphics[width=1.0\linewidth]{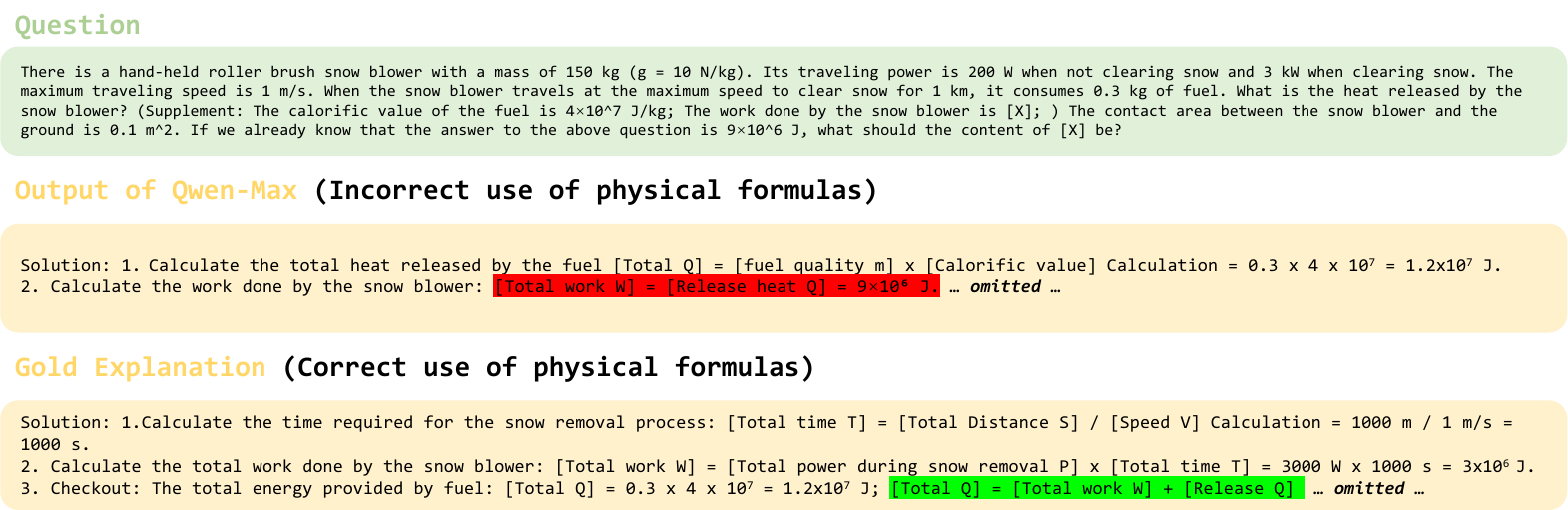}
    \caption{An error case caused by formula error.}
    \label{fig:subbb}
  \end{subfigure}
  \hfill
  \begin{subfigure}{\linewidth}
    \includegraphics[width=1.0\linewidth]{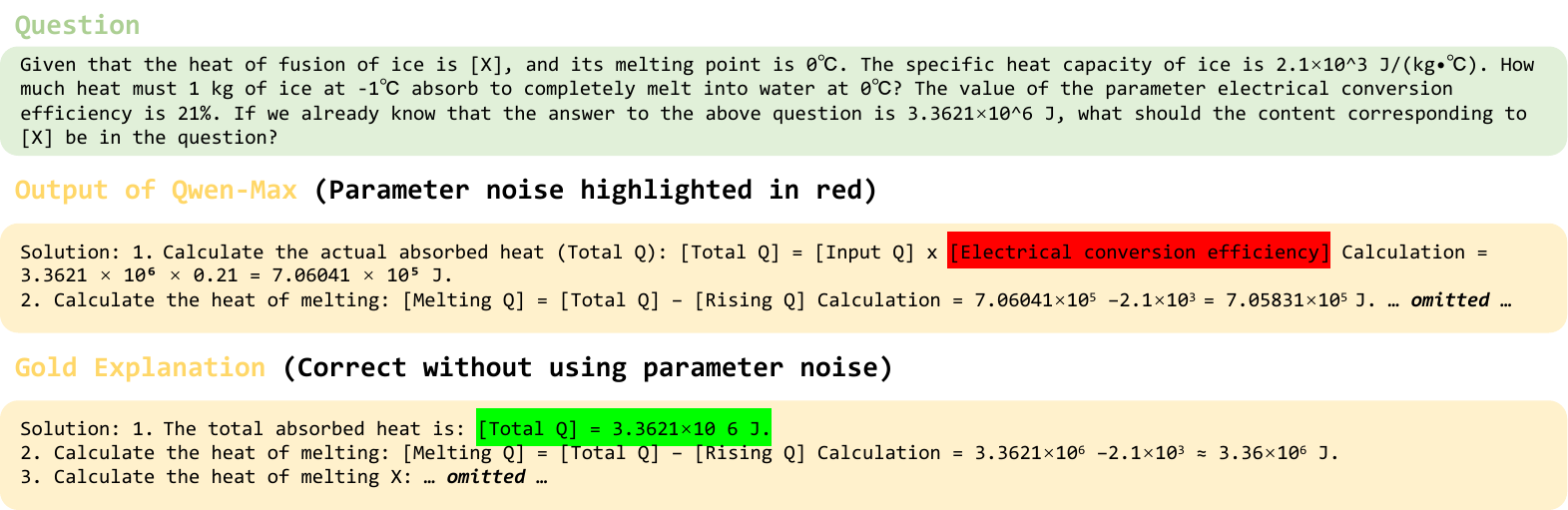}
    \caption{An error case caused by parameter noise.}
    \label{fig:subaa}
  \end{subfigure}
  \caption{Examples of errors. %
  }
  \label{fig:error_examples}
\end{figure*}

\datasetname\ posed challenges to existing models in terms of formula selection and numerical calculation, including unit calculation and arithmetic calculation. As illustrated in Figure~\ref{fig:subbb}, the incorrect use of formulas was the main type of error. Another type of error was parameter noise, as illustrated in Figure~\ref{fig:subaa}. In terms of the performance of the models, we found that the models that achieved higher accuracy were less affected by the parameter noise injected into the questions (Section~\ref{sec:question_alteration}). For example, among the questions to which DeepSeek-R1 did not provide a correct answer, approximately~14\% were influenced by the parameter noise. This indicated that our question alteration imposed further requirements on the model's ability to correctly apply formulas and perform accurate reasoning.

\end{document}